\documentclass[sigconf]{acmart}

\AtBeginDocument{%
  }

\usepackage{xcolor}
\usepackage[skip=3pt]{caption}

\usepackage{booktabs,multirow,makecell,adjustbox,array,colortbl}
\usepackage{xspace}
\usepackage{amsmath}
\usepackage{pifont}
\usepackage{wasysym}
\usepackage{enumitem}
\usepackage{tcolorbox}
\usepackage{booktabs,multirow,makecell,adjustbox,threeparttable,wasysym,pifont,appendix}

\newcommand{\dataset}{ClarifyMT-Bench\xspace}
\newcommand{\model}{ClarifyAgent\xspace}
\newcommand{\cmark}{\textcolor{green!60!black}{\ding{51}}}  % 绿色 ✓
\newcommand{\xmark}{\textcolor{red}{\ding{55}}}  

\definecolor{lightblue}{RGB}{230,245,255}
\definecolor{lightgreen}{RGB}{234,247,234}
\definecolor{lightyellow}{RGB}{255,249,225}
\definecolor{lightgray}{gray}{0.95}

\definecolor{darkred}{HTML}{8B0000}
\definecolor{lightgray}{gray}{0.8} % 柔和的浅灰背景

\newcommand{\contradictory}[1]{\colorbox{darkred!35}{{#1}}}

\newcommand{\vague}[1]{\colorbox{gray!50}{#1}}

\acmConference[]{}{}{}
\renewcommand\footnotetextcopyrightpermission[1]{} 

\title{ClarifyMT-Bench: Benchmarking and Improving Multi-Turn Clarification for Conversational Large Language Models}

\author{Sichun Luo$^1$, Yi Huang$^2$, Mukai Li$^1$, Shichang Meng$^3$, Fengyuan Liu$^1$, Zefa Hu$^2$, Junlan Feng$^2$, Qi Liu$^1$\\
$^1$The University of Hong Kong \quad $^2$JIUTIAN Research, China Mobile \quad $^3$CityUHK\\
\texttt{sichunluo2@gmail.com}\\ $ $
}

\begin{document}

%% ================= ABSTRACT =================
\begin{abstract}

Large language models (LLMs) are increasingly deployed as conversational assistants in open-domain, multi-turn settings, where users often provide incomplete or ambiguous information. However, existing LLM-focused clarification benchmarks primarily assume single-turn interactions or cooperative users, limiting their ability to evaluate clarification behavior in realistic settings. We introduce \textbf{\dataset}, a benchmark for multi-turn clarification grounded in a five-dimensional ambiguity taxonomy and a set of six behaviorally diverse simulated user personas. Through a hybrid LLM–human pipeline, we construct 6,120 multi-turn dialogues capturing diverse ambiguity sources and interaction patterns. Evaluating ten representative LLMs uncovers a consistent under-clarification bias: LLMs tend to answer prematurely, and performance degrades as dialogue depth increases. To mitigate this, we propose \textbf{\model}, an agentic approach that decomposes clarification into perception, forecasting, tracking, and planning, substantially improving robustness across ambiguity conditions. \dataset establishes a reproducible foundation for studying when LLMs should ask, when they should answer, and how to navigate ambiguity in real-world human–LLM interactions. 
% To facilitate future research, we release our code and dataset publicly at \url{https://anonymous.4open.science/r/ClarifyMT-Bench-635E}.
\end{abstract}

% \keywords{Clarifying Question, Ambiguity Type, Multi-turn Dialogue, LLM Evaluation}

\maketitle
% \pagestyle{plain}

%% ================= 1. INTRO =================
% \vspace{-0.1in}
\section{Introduction}

\begin{figure}[t]
    \centering
\includegraphics[width=0.43\textwidth]{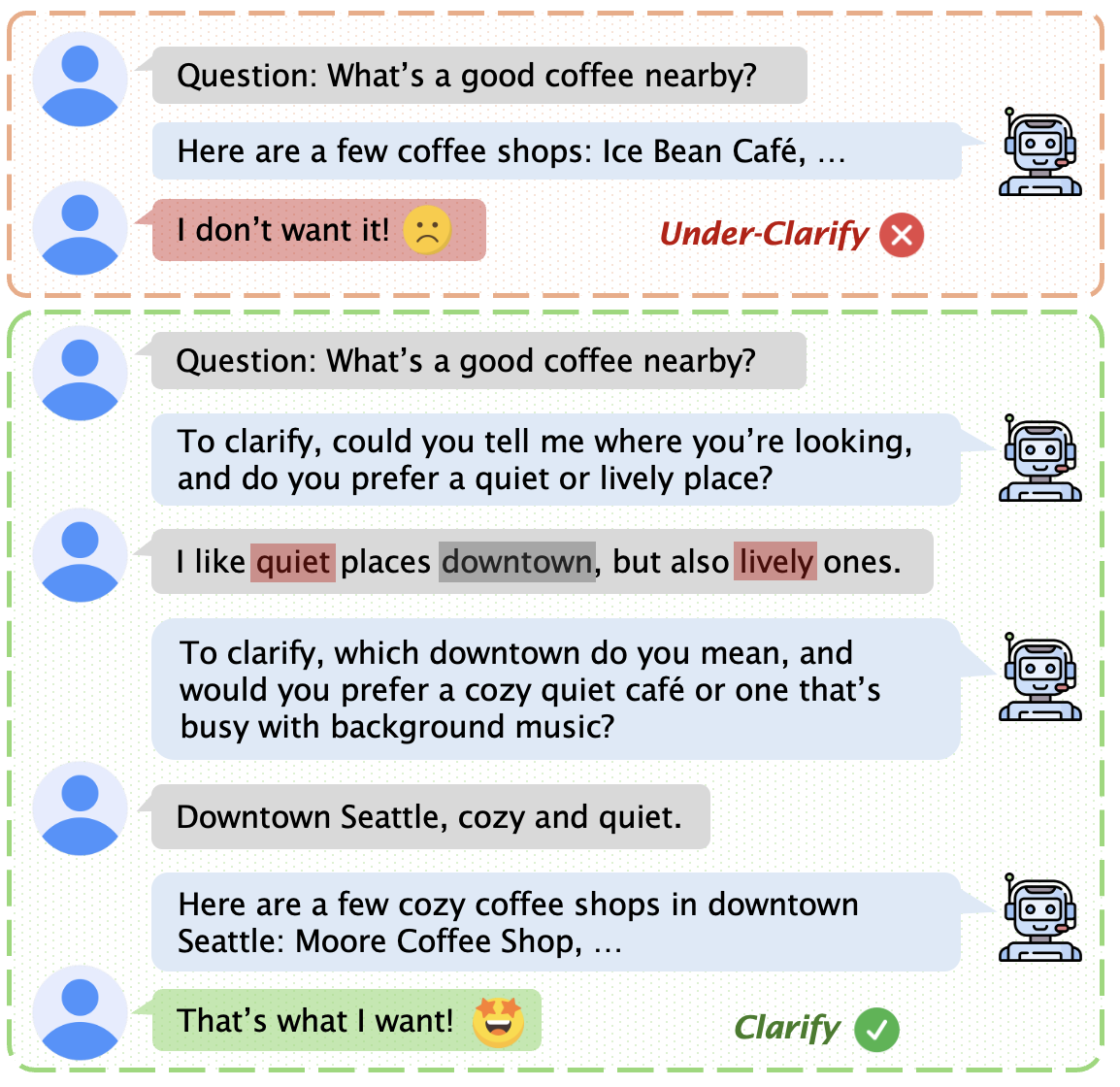}
    \caption{Illustration of clarification in user–LLM interactions. The upper example shows an under-clarified response that fails to capture user intent, while the lower example demonstrates effective clarification through follow-up questions, leading to user satisfaction. The user response may contain \contradictory{contradictory} or \vague{vague} information, labeled in red and gray respectively.
    % An illustration of multi turn dialogue clarification. llm need to repeatedly decide to ask or answer, facing noisy user that may contain Contradictory and partial vague information.
    }
\label{fig:demo}

\vspace{-0.05in}
\end{figure}

Large language models (LLMs) have rapidly become the foundation of modern conversational systems \cite{lo2023impact,wu2023brief,chatgpt2025}. Yet in real-world interactions, user inputs are rarely complete or unambiguous: users often omit key details, provide vague descriptions, contradict earlier statements, or respond off-topic~\cite{song2007identifying,mao2025prompts}. Since modern LLMs are primarily optimized for helpfulness and fluency~\cite{ouyang2022training,wang2023aligning}, they tend to generate confident answers even when the user intent is underspecified. Such premature answering behavior can lead to misleading or unsafe outputs, disproportionately affecting users with limited domain knowledge or digital literacy~\cite{nie2020adversarial,huang2024trustllm}.

Clarification offers a principled alternative: rather than inferring user intent from underspecified queries, a system should proactively ask targeted follow-up questions until the necessary information is obtained \cite{deng2023prompting,tang2025clarifying}. However, clarification in open-domain, multi-turn dialogue is inherently challenging, as user responses may be vague, contradictory, or off-focus, as illustrated in Figure~\ref{fig:demo}. Recent work has begun to explore clarification-oriented dialogue systems. CLAMBER~\cite{zhang2024clamber} evaluates ambiguity detection and single-turn clarifying question generation; ClariMM~\cite{ramezan2025multi} incorporates multimodal signals for disambiguation; and ClarQ-LLM~\cite{gan2024clarq} studies clarification in task-oriented domains. 
Despite these advances, existing efforts remain limited in scope and do not fully address clarification under noisy, unconstrained, and multi-turn interaction patterns.

% Yet these benchmarks typically assume single-turn interactions, cooperative user behavior, or constrained domains, limiting their ability to reflect the noisy and non-deterministic nature of real-world human--AI interactions. 

Existing benchmarks exhibit several key limitations: they typically assume single-turn interactions, cooperative user behavior, or narrowly scoped tasks, limiting their ability to capture the noisy and non-deterministic nature of real-world human-LLM interactions. Moreover, existing evaluations rarely assess when an LLM should ask, what it should ask, when it should stop asking, or how it should remain robust under vague, contradictory, or factually incorrect user feedback, which are the core challenges real-world conversational settings. This motivates a central question:
\textit{\textbf{Can LLMs effectively balance clarification and answering under complex, noisy multi-turn interactions?}}
% \end{quote}

To address this question, we introduce \textbf{\dataset}, a multi-turn clarification benchmark designed to evaluate an LLM’s ability to dynamically choose between {asking} and {answering}. Our benchmark is grounded in a five-dimensional ambiguity taxonomy that spans linguistic, intent, contextual, epistemic, and interactional ambiguity, providing a principled foundation for controlled multi-turn evaluation. We construct the dataset using a hybrid LLM–human pipeline that generates diverse ambiguous queries, filters and refines them for correctness and clarity, and expands them into multi-turn dialogues. To model realistic conversational uncertainty, we further incorporate a user-behavior simulator encompassing six response types
via LLM prompting and validated by human annotators for quality and diversity.
A comparison of \dataset with prior benchmarks is provided in Table~\ref{tab:bench-compare}.

\begin{table}[!t]
\centering
\caption{Comparison with related datasets and benchmarks.}
\label{tab:bench-compare}
\begin{threeparttable}
\begin{adjustbox}{width=0.95\linewidth}
\begin{tabular}{l|ccccc}
\toprule
\makecell[l]{Dataset / Benchmark} &
\makecell{Clarify} &
\makecell{Multi-\\Turn} &
\makecell{Open-\\Domain} &
\makecell{LLM-\\focused\\Eval} &
\makecell{Noisy\\User} \\
\midrule
ClariQ  \cite{aliannejadi2020convai3}  & \cmark & \xmark & \cmark & \xmark & \xmark \\
CLAMBER~\cite{zhang2024clamber}          & \cmark & \xmark & \cmark & \cmark & \xmark \\
ClarQ-LLM~\cite{gan2024clarq}            & \cmark & \cmark & \xmark & \cmark & \xmark \\
ClariMM \cite{ramezan2025multi}          & \cmark & \cmark & \xmark & \cmark & \xmark \\
% ShARC / OR-ShARC \cite{saeidi2018sharc,gao2021orsharc} & \cmark & \cmark & \xmark & \xmark & \xmark \\
% TREC CAsT~\cite{dalton2020cast}          & \xmark & \cmark & \xmark & \xmark & \xmark \\
% MIMICS / MIMICS-Duo~\cite{zamani2020mimics,tavakoli2022mimicsduo} & \cmark & \cmark & \xmark & \xmark & \xmark \\
\midrule
\textbf{Our \dataset}                    & \cmark & \cmark & \cmark & \cmark & \cmark \\
\bottomrule
\end{tabular}
\end{adjustbox}
\end{threeparttable}
\end{table}

We evaluate ten leading LLMs spanning six model families. Across all settings, we observe a consistent under-clarification bias: most LLMs prefer to answer prematurely rather than engage in sufficient clarification, especially as dialogue depth increases. This finding highlights an important gap between current alignment objectives and robust, responsible conversational behavior.

To address this challenge, we propose \textbf{\model}, an agent-based framework that formulates multi-turn clarification as a structured reasoning process.
Beyond the ask–answer decision task, we introduce a new user persona inference task.
Specifically, upon receiving a user query, the Perceiver extracts task-relevant information and identifies potential ambiguity, while the Forecaster infers the user persona to anticipate the user’s behavioral tendencies.
The Tracker then updates the state representing unresolved ambiguity slots.
Finally, the Planner integrates signals from all components and decides whether to continue clarifying or proceed to answering, with the selected action executed by the Output module.
Experimental results validate the effectiveness of \model, demonstrating substantial improvements in clarification performance.

% \textit{Contributions.}
% In summary, this work makes the following contributions:
In a nutshell, our contributions are threefold.
\begin{itemize}[leftmargin=1.2em]

\item \textbf{Benchmark.} We present \dataset, 
to our best knowledge, the first benchmark that jointly evaluates multi-turn, open-domain clarification under noisy user behavior. 
Grounded in a five-dimensional ambiguity taxonomy and six behaviorally diverse user personas, 
\dataset{} contains 6{,}120 multi-turn interactions, enabling controlled evaluation under diverse conversational uncertainty.

\item \textbf{Analysis.} Through extensive experiments on ten representative LLMs, we identify a consistent under-clarification bias and quantify robustness disparities across ambiguity types, user persons, and dialogue depth. These findings highlight key limitations in current conversational LLMs’ ability to balance asking and answering under uncertainty.

\item \textbf{Method.} We introduce \model, an agentic framework that decomposes clarification into perception, forecasting, tracking, and planning. 
By introducing extra user persona inference task,
\model{} produces more robust multi-turn clarification behavior, serving as a strong baseline for future research.
\end{itemize}

\begin{table*}[t]
\centering
\caption{Ambiguity Taxonomy in the era of LLMs, illustrating five major dimensions and representative subtypes with examples. Categories are organized along a continuum from linguistic form to social interaction. }
\small
\renewcommand{\arraystretch}{1.2}
\resizebox{\textwidth}{!}{%
\begin{tabular}
{p{3cm} p{3.3cm} p{6.5cm} p{6.5cm}}
\toprule
\textbf{Category} & \textbf{Subtype} & \textbf{Description}& \textbf{Example} \\
\midrule

\multirow{3}{*}{\textbf{Linguistic Ambiguity}} 
& \textit{Lexical Ambiguity} & Word has multiple meanings. & Please tell me about the seal. \\
& \textit{Syntactic Ambiguity} & Sentence allows multiple parses. & List movies from the 1990s starring actors from Canada. \\
& \textit{Semantic Ambiguity} & Unclear semantic role or criteria. & Is New York the largest city? \\

\midrule

\multirow{3}{*}{\textbf{Intent Ambiguity}} 
& \textit{Goal Ambiguity} & The user’s goal is vague or incomplete. & Help me write a report. \\
& \textit{Scope Ambiguity} & The intended task scope is unclear. & Tell me about quantum computing.  \\
& \textit{Intent Conflict Ambiguity} & The user expresses incompatible goals. & Summarize 'War and Peace' without omitting anything. \\

\midrule

\multirow{3}{*}{\textbf{Contextual Ambiguity}} 
& \textit{Entity Ambiguity} & Multiple possible referents exist for a term or name. & Who is the real Spider-Man? \\
& \textit{Spatial Ambiguity} & The location is unspecified or underspecified. & Tell me how to reach London. \\
& \textit{Temporal  Ambiguity} & The time is unspecified or underspecified. & I need the weather forecast for New York. \\
% & \textit{Anaphoric Ambiguity} & Unclear or underspecified pronoun reference & Was it better than the last one? \\

\midrule

\multirow{3}{*}{\textbf{Epistemic Ambiguity}} 
& \textit{Knowledge Gap Ambiguity} & The user assumes shared prior knowledge or context. & You remember the new update, right? \\
& \textit{Unfamiliarity Ambiguity} & The query involves entities or facts unknown to the model. & Find the price of the Samsung Chromecast. \\
% & \textit{Model epistemic humility} & The query challenges or probes the model’s self-awareness or limits. & You must know everything about this topic, don’t you? \\

& \textit{Value Ambiguity} & Subjective or evaluative terms without clear criteria. &Recommend a good movie. \\

\midrule

\multirow{4}{*}{\textbf{Interactional Ambiguity}} 
& \textit{Partial / Vague reply} & The user provides uncertain or imprecise feedback. & Sort of, I guess. \\
& \textit{Factually Wrong Reply} & The user provides information that is clearly incorrect. & Paris is the capital of Germany. \\
& \textit{Contradictory Reply} & The user expresses conflicting statements or attitudes. & It’s urgent. No rush actually. \\
& \textit{Off-focus Reply} & The user diverts  the intended topic or clarification. & Let’s talk about something else. \\

\bottomrule
\end{tabular}
}
\label{tab:ambiguity-taxonomy}
\end{table*}

% in preamble:
% \usepackage{booktabs,makecell,adjustbox,array}

%% ================= 2. RELATED =================
\section{Related Work}

\textit{Multi-turn Dialogue.}
Multi-turn dialogue better reflects real-world conversational settings and plays a central role in user experience~\cite{zheng2023judging,bai2024mt}.
However, recent studies show that LLMs still struggle with state tracking, long-context reasoning, and grounding under evolving dialogue histories~\cite{yi2024survey,li2025beyond,laban2025llms}.
When instructions involve pronouns, ellipses, or cross-turn dependencies, models frequently fail to maintain consistency or follow user intent~\cite{chenlearning}.
Such limitations directly affect clarification, where models must track ambiguous references, integrate new information, and determine when clarification is necessary as the dialogue progresses.

\textit{LLM-oriented Clarification Dataset and Benchmark.}
Traditional clarification methods in IR and NLP treat ambiguity as a static property of the query, resolving lexical, syntactic, or topical underspecification through query expansion, disambiguation, or slot filling~\cite{krovetz1992lexical,navigli2009word}.
In contrast, LLM-based assistants must reason under open-ended generation, hallucination risk, epistemic uncertainty, and diverse user preferences, making the ask–or–answer decision itself central.
Several recent benchmarks study this direction: \textsc{CLAMBER}~\cite{zhang2024clamber} evaluates clarifying question quality; ClariMM~\cite{ramezan2025multi} extends the task to multimodal inputs; and ClarQ-LLM~\cite{gan2024clarq} examines multi-turn clarification in task-oriented settings.
While valuable, these resources typically do not jointly evaluate stop decisions, operate in limited domains, or account for noisy and inconsistent users, leaving open how to assess clarification policies under realistic ambiguity and imperfect user feedback.

\textit{Clarification Methods for LLMs.}
Prompting-based approaches have been explored to elicit better clarification behavior.
Deng et al.~\cite{deng2023prompting} and Lee et al.~\cite{lee2023asking} investigate prompting strategies for asking clarifying questions.
Moreover, Zhang et al.~\cite{zhang2025clarify} introduce a task-agnostic framework for detecting ambiguity, selecting effective clarifying questions, and integrating new information.
Additionally, AT-CoT~\cite{tang2025clarifying} improves clarification by first identifying ambiguity type and then generating targeted questions.
Although effective in single-turn or cooperative settings, these methods do not directly generalize to multi-turn interactions with noisy, inconsistent, or adversarial users, where clarify-or-answer decisions must be made sequentially under uncertainty.

\textit{User Simulation with LLMs.}
User simulators have long been used to evaluate interactive systems, from task-oriented dialogue~\cite{sun2023metaphorical} to interactive IR~\cite{maxwell2016simulating}. However, many classical simulators assume cooperative, truthful, and stationary users, limiting their ability to reflect the variability and noise present in real interactions \cite{wang2024depth}. Recent LLM-based simulators introduce richer behaviors and improved linguistic naturalness, yet they frequently emphasize coherent and goal-directed responses, which may overlook contradictory, off-focus, or factually incorrect replies \cite{sekulic2024reliable,luo2024duetsim}.
Capturing such behaviors is essential for robust clarification evaluation, where systems must determine when to ask, what to ask, and when to stop under imperfect and unpredictable user feedback.

\textit{Limitations of Existing Work.}
Existing LLM-oriented clarification benchmarks have advanced the study of ambiguity resolution, but they often restrict domain coverage or interaction types, limiting their applicability to open-domain conversational settings.
Moreover, current ambiguity taxonomies provide limited expressiveness for modeling the multi-faceted sources of ambiguity that arise in LLM-era interactions.
Most prior work further assumes cooperative or noise-free user feedback, under-emphasizing robustness to vague, off-topic, contradictory, or factually incorrect responses.
In addition, existing benchmarks rarely evaluate state-of-the-art LLMs under such challenging conditions.
These limitations highlight the need for a more comprehensive benchmark capable of evaluating clarification policies under realistic, multi-turn, and noise-prone human–LLM interactions.

% moreover, some works do not use sota llms

% though these 

% these definition of Ambiguity is not focused on llm era and are narrow scoped.

% Our benchmark addresses these gaps by 

% (a) formalizing \emph{Required Slot Completion} (RSC) as a deterministic sufficiency criterion to label \emph{Under-Clarify}, \emph{Over-Clarify}, and \emph{Irrelevant-Clarify}; (b) covering general web-style queries beyond a single vertical; (c) instantiating a user simulator with orthogonal dimensions—information coverage, truthfulness, self-consistency, cooperativeness, and specificity—to yield six reusable behavior types; and (d) adding an auxiliary \emph{Ambiguity Forecasting \& User-Type Inference} task to calibrate ask rates and adapt strategies online. Together, these design choices enable rigorous, reproducible evaluation of LLM clarification policies in realistic multi-turn settings.

%% ================= 3. TASK =================

\section{Taxonomy and Task}

\subsection{Ambiguity Taxonomy in the Era of LLMs}
\label{sec:clarification-types}

Ambiguity is an inherent property of human communication, and prior work typically categorizes it into syntactic, semantic, and contextual varieties~\cite{zhang2024clamber}. However, LLM-based assistants encounter a broader range of ambiguity sources that extend beyond traditional linguistic formulations. We therefore propose a five-dimensional taxonomy spanning {linguistic}, {intent}, {contextual}, {epistemic}, and {interactional} ambiguity. This taxonomy reflects a progression from surface-level linguistic form to interaction-level behavior, consistent with cognitive models of communication that link symbols, intentions, shared context, and collaborative action~\cite{clark1992arenas,levinson2000presumptive}.
Our five-dimensional taxonomy is defined as follows:

\begin{itemize}[leftmargin=1.2em]

    \item \textit{Linguistic Ambiguity.}  
    Ambiguity arising from lexical, structural, or semantic indeterminacy in the user's utterance.

    \item \textit{Intent Ambiguity.}  
    Uncertainty about what the user wants the system to accomplish, including unclear goal, task scope, or potentially conflicting intent.

    \item \textit{Contextual Ambiguity.}  
    Ambiguity caused by under-specified or shifting references to entities, locations, time, or discourse context.

    \item \textit{Epistemic Ambiguity.}  
    Ambiguity about the knowledge shared between the user and the model, such as assumptions about background knowledge, unfamiliar entities, or subjective evaluative criteria.

    \item \textit{Interactional Ambiguity.}  
    Ambiguity introduced by the interaction itself when user replies are vague, off-focus, contradictory, or factually incorrect, making it unclear how the conversation should proceed.

\end{itemize}

This five-dimensional taxonomy highlights the multifaceted nature of ambiguity in LLM-mediated communication and provides a conceptual foundation for evaluating models' ability to detect, diagnose, and adapt to diverse uncertainty sources. Representative subtypes and examples are provided in Table~\ref{tab:ambiguity-taxonomy}.

\subsection{Task Formulation}

We view multi-turn clarification as a sequential decision-making problem. 
Given an under-specified user query $q_0$, the goal of the dialogue system is to acquire the minimal additional information necessary to produce a well-specified and complete answer.

Conceptually, we assume a set of ambiguity-relevant slots $S = \{s_1, \dots, s_n\}$ (\textit{e.g.}, destination, budget, time), and a subset $S^* \subseteq S$ denoting the \emph{required slots} that must be resolved for the query. At turn $t$, the dialogue state can be represented as
\begin{align}
    x_t = [f_t(s_1), \dots, f_t(s_n)],
\end{align}
where each slot is in one of three abstract states:
\begin{align}
    f_t(s_i) \in \{\texttt{unfilled}, \texttt{filled}, \texttt{conflict}\}.
\end{align}
An ideal clarification policy would choose an action $a_t \in \{\texttt{Clarify}, \\\texttt{Answer}\}$ based on $x_t$, and would stop asking once all required slots are filled and no conflicts remain:
\begin{align}
    \text{Stop if} \quad \Big( \forall s \in S^*, f_t(s) = \texttt{filled} \Big) \land \Big( \nexists s' \in S, f_t(s') = \texttt{conflict} \Big).
    \label{eq:stopping}
\end{align}

In practice, we do not explicitly annotate slots or $S^*$ for each instance, due to the high cost and ambiguity of manually identifying required information for open-domain queries.
Instead, each dialogue turn in \dataset{} is associated with a \emph{reference action} $y_t \in \{\texttt{Clarify}, \texttt{Answer}\}$ derived from our construction pipeline. 
Given the dialogue prefix up to turn $t$, the model predicts an action $\hat{a}_t$, and we evaluate its decisions as follows:
\begin{itemize}[leftmargin=1.2em]
    \item \emph{Under-Clarify}: $\hat{a}_t = \texttt{Answer}$ while $y_t = \texttt{Clarify}$, meaning the model answers prematurely.
    \item \emph{Over-Clarify}: $\hat{a}_t = \texttt{Clarify}$ while $y_t = \texttt{Answer}$, meaning the model asks when an answer is already warranted.
\end{itemize}

We primarily report overall decision accuracy, \textit{i.e.}, $\mathbb{I}[\hat{a}_t = y_t]$ averaged over all turns and instances, and use the under-/over-clarification categories in our qualitative analysis.

% \begin{figure}[t]
%     \centering
% \includegraphics[width=0.45\textwidth]{z3.png}
%     \caption{An illustration of multi turn dia. llm meet more challenge in multi turn dia.}
% \label{fig:pipeline}

% \vspace{-0.1in}
% \end{figure}

% \subsection{Answer Acceptability}
% Each episode includes \textit{answer checks}: keyword sets, schema fields, or logic predicates. For safety-adjacent tasks, we require presence of warnings/disclaimers. For stylistic tasks (e.g., “movie quotes”), we check format (count, de-duplication) rather than content truthfulness.

%% ================= 4. DATA =================

\begin{figure}[!t]
    \centering
\includegraphics[width=0.49\textwidth]{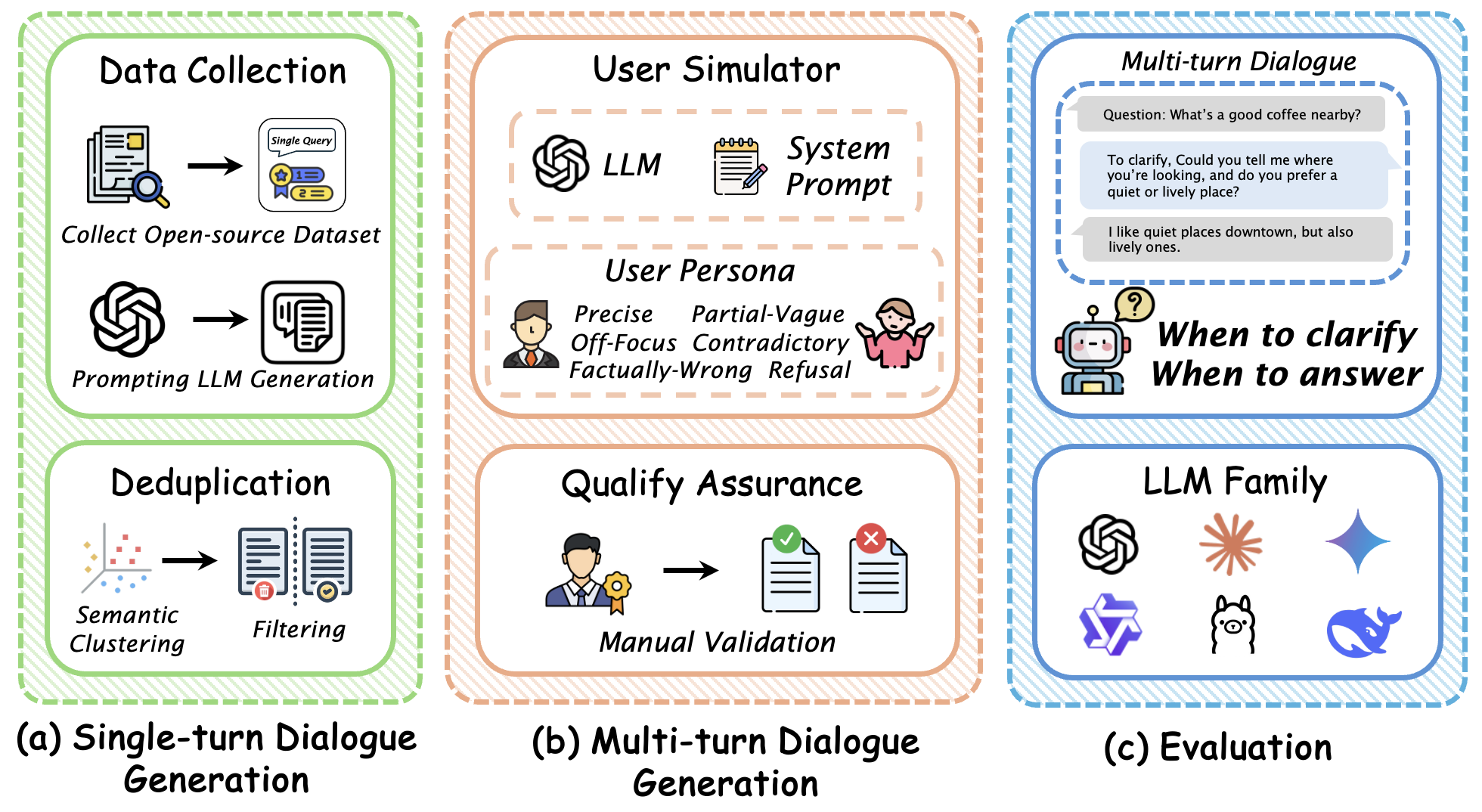}
\vspace{0.01in}
    \caption{The pipeline of dataset construction.}
\label{fig:pipeline}

% \vspace{-0.1in}
\end{figure}

\section{Dataset Construction}

As illustrated in Figure~\ref{fig:pipeline}, the construction of \dataset~consists of two stages:  
(1) \emph{single-turn dialogue generation}, where we collect and synthesize ambiguous queries paired with clarifying questions; and  
(2) \emph{multi-turn dialogue expansion}, where we simulate user responses of different behavior types to extend each instance into a multi-turn conversation.  
This pipeline yields a benchmark that reflects the complexity and noise patterns observed in real-world clarification-oriented dialogue.

\subsection{Single-Turn Dialogue Generation}

\textit{Data Collection.}
We first construct an initial pool of ambiguous user queries and their corresponding clarifying questions.  
To ensure broad coverage, we prompt diverse LLMs (\textit{e.g.}, GPT-4.1, DeepSeek-V3) to generate ambiguous queries along with clarifying questions aligned with the taxonomy in Section~\ref{sec:clarification-types}.  
Prompt templates and instructions are provided in Figure~\ref{fig:prompt}.

Formally, let $\mathcal{M}$ denote an LLM and $p$ a prompt template containing task instructions and few-shot exemplars.  
The model outputs an ambiguous query $q_{\text{amb}}$ paired with a clarifying question $q_{\text{clar}}$:
\begin{equation}
(q_{\text{amb}}, q_{\text{clar}}) = \mathcal{M}(p; \theta),
\end{equation}
where $\theta$ represents model parameters.  
We repeat this process across multiple models and prompt variants to increase lexical, stylistic, and semantic diversity.

In addition, we leverage samples from existing clarification datasets, such as \textsc{CLAMBER} \cite{zhang2024clamber}, to improve diversity.
% To improve category balance, we prompt LLMs to generate supplemental clarifying questions conditioned on specific clarification types:
% \begin{equation}
% q_{\text{clar}}^{(t)} = \mathcal{M}(q_{\text{amb}}, c_t),
% \end{equation}
% where $c_t$ denotes a predefined clarification type (e.g., \emph{attribute}, \emph{intent}, \emph{entity}).  
% This augmentation ensures comprehensive representation across ambiguity categories.

\textit{Deduplication.}
To improve quality and prevent redundancy, we perform semantic-level deduplication.  
To be specific, each query $q_i$ is embedded using a pretrained encoder~$\mathcal{E}$:
\begin{equation}
\mathbf{v}_i = \mathcal{E}(q_i),
\end{equation}
and cosine similarity is computed for each pair:
\begin{equation}
S_{ij} = \frac{\mathbf{v}_i^\top \mathbf{v}_j}{\|\mathbf{v}_i\| \, \|\mathbf{v}_j\|}.
\end{equation}
Queries with similarity $S_{ij} > \tau_{\text{sem}}$ are grouped into cluster~$\mathcal{C}_k$.  
From each cluster, we retain a single representative:
\begin{equation}
q_k^{*} = \arg\max_{q_i \in \mathcal{C}_k} \text{Quality}(q_i),
\end{equation}
where $\text{Quality}(q_i)$ reflects fluency and informativeness assessed through human annotation.  
We use \texttt{all-MiniLM-L6-v2}\footnote{https://huggingface.co/sentence-transformers/all-MiniLM-L6-v2}
as the encoder and set $\tau_{\text{sem}} = 0.7$ based on a small pilot study to balance recall and diversity.  
This procedure removes semantic duplicates and enhances dataset variety.

\subsection{Multi-Turn Dialogue Generation}

\textit{User Simulator.}
To emulate realistic human variability, we model user behavior along five dimensions: information coverage, truthfulness,
self-consistency, cooperativeness, and specificity.
These dimensions reflect key factors that influence clarification difficulty and commonly vary in real-world interactions.  
Based on them, we define six canonical user personas:

\begin{itemize}[leftmargin=1.2em]
    \item \textit{Precise}: provides specific, accurate, and fully relevant information that directly fills the target slot.
    \item \textit{Partial–Vague}: offers partially relevant but underspecified or indecisive responses with low specificity.
    \item \textit{Off–Focus}: replies with off-topic or tangential content that does not address the requested slot.
    \item \textit{Contradictory}: gives internally or cross-turn inconsistent responses that hinder stable intent inference.
    \item \textit{Factually–Wrong}: provides incorrect factual information. 
    \item \textit{Refusal}: declines to clarify, avoids answering, or explicitly urges the model to proceed without clarification.
\end{itemize}

This structured simulator supports controlled evaluation across cooperative, noisy, and adversarial user behaviors.  
{Precise} and {Refusal} users typically provide enough information or explicitly decline to provide more, permitting a direct answer, whereas the other four personas generally require additional clarification.
Representative examples of user replies are shown in Table~\ref{tab:user-types-travel}.

\textit{Manual Validation.}
Following automatic generation, we conduct a human quality check.  
Two annotators with prior NLP data-curation experience independently review each ambiguous query – clarification pair.  
Instances that are incoherent, ungrammatical, mislabeled, or inconsistent with the intended ambiguity category are removed.  
Disagreements are resolved through discussion until consensus is reached.  
This verification step ensures that all retained samples are linguistically sound and faithfully aligned with their target ambiguity types.

\textit{Human Annotation Reliability.}
To evaluate the reliability of manual annotations, two annotators independently performed a binary filtering decision (retain or remove) on 10\% of the randomly sampled instances. We compute Cohen's Kappa ($\kappa$) \citep{cohen1960coefficient}, which adjusts for chance agreement. The observed agreement is $P_o = 0.8627$, and the resulting Kappa is $\kappa = 0.5980$, indicating moderate to substantial agreement according to the Landis \& Koch scale \citep{landis1977measurement}. These results demonstrate that the annotation guideline is clear and that the filtering quality is consistent across annotators.

\subsection{Data Statistics}
Our dataset covers 12 fine-grained ambiguity subtypes, grouped into four broad categories: linguistic, intent, contextual, and epistemic ambiguity, which characterize the source of ambiguity in the initial query.  
We generate 85 single-turn instances for each subtype, resulting in 1,020 ambiguous query–clarification pairs.
To capture interaction-level variability in user responses, each single-turn instance is further expanded into six multi-turn dialogues, corresponding to six simulated user personas.  
For the Precise and Refusal user personas, we construct two-turn conversations, while we generate three-turn conversations for the remaining four personas.
We limit each simulated dialogue to two or three turns to keep the evaluation focused and controllable.
Longer dialogues are certainly possible in real-world settings, but extending them synthetically risks accumulating distributional artifacts (\textit{e.g.}, unnatural conversational drift) and compounding model errors from the user simulator.
This design produces a total of \textbf{6,120} multi-turn dialogues, with an average of \textbf{2.67} turns per dialogue.
As shown in Figure~\ref{fig:dia_length}, the average token length per dialogue is \textbf{66.1} tokens.
The data source distribution is provided in Appendix~\ref{sec:adddata}.

% We report dataset statistics in Table~\ref{tab:dataset-stats}, including the number of instances per ambiguity category, average tokens per user query, average length of clarifying questions, and average number of dialogue turns.  
% The dataset maintains a balanced distribution across ambiguity categories and user types, providing comprehensive coverage for both model training and evaluation.

\section{Evaluation}

\subsection{Evaluation Setup}

% \subsection{Evaluation Protocol and Metrics}
% We report:
% \begin{itemize}[leftmargin=1.2em]
% \item \textbf{Acc}: final answer acceptability.
% \item \textbf{UCR}: Under-Clarify rate (answered before \rsc).
% \item \textbf{OCR}: Over-Clarify rate (asked after \rsc).
% \item \textbf{ICR}: Irrelevant-Clarify rate (asked off-target, not resolving conflict).
% \item \textbf{TTS}: Turns-to-Sufficiency (turns until \rsc\ satisfied).
% \end{itemize}
% We macro-average across user types and report per-type robustness (Acc@P/PV/CT/FA/RF) and \emph{Recovery@CT/FA} (fraction of episodes where conflicts are resolved before answering).

%% ================= 6. AF-UTI =================

% {User-Type Inference.}
% We introduce an additional task, User-Type Inference.
% After 1--2 user turns, the agent predicts the current user type (P/PV/OF/CT/FA/RF). Metrics: macro-F1 and downstream deltas in UCR/OCR/ICR/TTS when strategies adapt based on the inferred type.

% We evaluate accuracy.

\textit{Tasks and Metrics.}
We evaluate models on two complementary tasks that together assess both clarification capability and dialogue quality.
(1) Multi-turn ambiguous query clarification.
Given an underspecified user query, a model must decide whether to ask a clarifying question or directly provide an answer.
We measure clarification accuracy, defined as the proportion of correctly selected actions over all dialogue turns.
We further report performance by user persona and ambiguity subtype to capture fine-grained behavioral differences.
All experiments are conducted on the full dataset, and we report the average accuracy.
(2) Clarifying question quality evaluation.
For each clarification turn, we evaluate the semantic quality of the generated question.
We adopt an LLM-as-a-Judge evaluation protocol~\cite{li2025generation} and manual evaluation to assess contextual appropriateness, relevance, and helpfulness.

% Specifiaclly, we randomly select 30 dialogues from each ambulityu subtype. we let GPT-4.1 to give a score from 
% This dual-metric evaluation captures both lexical alignment and pragmatic quality.

% Following prior work, we compute BERTScore to assess semantic similarity against reference clarifications, and 

\begin{table}[!t]
\centering
\caption{Model performance across six user personas: Precise (P), Partial-Vague (PV), Off-Focus (OF), Contradictory (CT), Factually-Wrong (FA), and Refusal (RF).
The best and worst results are labeled in green and red, respectively.
}
\label{tab:perf_user_type}
\begin{adjustbox}{width=\linewidth}
\begin{tabular}{lccccccc}
\toprule
\textbf{Model} & \textbf{P } & \textbf{PV } & \textbf{OF } & \textbf{CT } & \textbf{FA } & \textbf{RF } & \textbf{Avg. } \\
\midrule

GPT-4.1 
& 69.2 
& 81.9 
& 89.1 
& 82.0 
& 69.9 
& 40.7 
& 72.1 \\

o3  
& 55.9 
& 73.3 
& 76.8 
& 68.1 
& 54.7 
& 56.5 
& 64.2 \\

Gemini-2.5-Flash 
& 48.7 
& \cellcolor{green!20}95.5 
& 94.5 
& \cellcolor{green!20}97.4 
& 83.7 
& \cellcolor{red!20}14.5 
& 72.4 \\

Claude-Sonnet-4.5 
& \cellcolor{red!20}36.8 
& 94.8 
& \cellcolor{green!20}96.1 
& 91.9 
& \cellcolor{green!20}93.7 
& 35.6 
& 74.8 \\

Qwen-2.5-7B-It 
& \cellcolor{green!20}96.6 
& \cellcolor{red!20}36.5 
& \cellcolor{red!20}60.3 
& \cellcolor{red!20}42.4 
& \cellcolor{red!20}38.8 
& \cellcolor{green!20}72.9 
& \cellcolor{red!20}57.9 \\

Qwen-2.5-72B-It 
& 93.8 
& 67.7 
& 78.9 
& 72.5 
& 75.4 
& 59.8 
& 74.7 \\

Llama-3.1-8B-It 
& 48.3 
& 75.6 
& 87.5 
& 86.4 
& 84.8 
& 44.4 
& 71.2 \\

Llama-3.1-70B-It 
& 70.4 
& 85.1 
& 90.9 
& 88.7 
& 88.7 
& 39.0 
& \cellcolor{green!20}77.1 \\

DeepSeek-V3 
& 70.0 
& 85.0 
& 82.1 
& 85.6 
& 78.6 
& 31.5 
& 72.1 \\

DeepSeek-R1 
& 68.0 
& 77.4 
& 78.8 
& 73.7 
& 64.4 
& 49.1 
& 68.6 \\

\bottomrule
\end{tabular}
\end{adjustbox}
\vspace{-0.1in}
\end{table}

\textit{Evaluated Models.}
We conduct extensive experiments using representative models from six major LLM families, covering both closed-source and open-source models, including:
GPT-4.1~\cite{openai_gpt41},
o3~\cite{openai_gpto3},
Gemini-2.5-Flash \cite{comanici2025gemini},
Claude-Sonnet-4.5~\cite{claude},
Qwen-2.5~\cite{qwen2025qwen25technicalreport},
Llama-3.1~\cite{dubey2024llama},
DeepSeek-V3~\cite{liu2024deepseek},
and DeepSeek-R1~\cite{guo2025deepseek}.
These LLMs span a range of training paradigms, allowing for a comprehensive comparison of robustness and interactional behavior under ambiguity.
The implementation details are shown in Appendix \ref{sec:impl}.

% \textit{Implementation Details}
% \label{sec:impl}
% Unless otherwise stated, we set the decoding temperature to 0, \textit{i.e.}, greedy decoding.
% Due to resource constraints, Qwen-2.5-7B-Instruct and Llama-3.1-8B-Instruct are locally deployed and served via the vLLM framework \cite{kwon2023efficient}.
% OpenAI models are accessed through their official APIs, while the DeepSeek models are accessed through the \texttt{Bailian} platform.\footnote{\url{https://aliyun.com/product/bailian}}
% All other models are accessed via the \texttt{OpenRouter} API service.\footnote{\url{https://openrouter.ai/}}
% Prompt templates follow the general format used in prior clarification benchmarks~\cite{zhang2024clamber,gan2024clarq}, with minor adjustments to ensure consistency across different model families.
% The complete prompt template is provided in Figure~\ref{fig:prompt}.

\begin{table*}[!t]
\centering
\caption{Model performance by ambiguity {subtype} across dialogue turns.
Subtypes include: Lexical (Lex.), Syntactic (Syn.), Semantic (Sem.), Goal (Goal), 
Scope (Sco.), Intent Conflict (Con.), Entity (Ent.), Spatial (Spa.), Temporal (Tmp.), 
Knowledge Gap (Kno.), Unfamiliarity (Unf.), and Value (Val.) ambiguity. 
Avg.\ denotes the average accuracy across all subtypes.
The best and worst
results are labeled in green and red, respectively.
}
\vspace{0.07in}
\label{tab:main}
\begin{adjustbox}{width=0.95\linewidth}
\begin{tabular}{lcccccccccccccc}
\toprule
\textbf{Model} & \textbf{Lex.} & \textbf{Syn.} & \textbf{Sem.} & \textbf{Goal} & \textbf{Sco.} & \textbf{Con.} & \textbf{Ent.} & \textbf{Spa.} & \textbf{Tmp.} & \textbf{Kno.} & \textbf{Unf.} & \textbf{Val.} & \textbf{Avg.}\\

\midrule
\rowcolor{cyan!10} \multicolumn{14}{c}{\textit{\textbf{Turn = 1}}} \\ \midrule

GPT-4.1 & 78.8 & 67.1 & 84.7 & \cellcolor{green!20}100.0 & 63.5 & 87.1 & 62.4 & 61.2 & \cellcolor{green!20}100.0 & \cellcolor{green!20}100.0 & 70.6 & 91.8 & 77.7 \\
o3 & 90.6 & 57.6 & 49.4 & 97.6 & 28.2 & 77.6 & 74.1 & 64.7 & 77.6 & \cellcolor{green!20}100.0 & 77.6 & 67.1 & 71.8 \\
Gemini-2.5-Flash & 88.2 & \cellcolor{green!20}100.0 & 94.1 & \cellcolor{green!20}100.0 & \cellcolor{green!20}85.9 & \cellcolor{green!20}90.6 & \cellcolor{green!20}77.6 & 69.4 & 83.5 & 97.6 & 87.1 & \cellcolor{green!20}98.8 & \cellcolor{green!20}89.4 \\
Claude-Sonnet-4.5 & 82.4 & 74.1 & 71.8 & 97.6 & 27.1 & 76.5 & 64.7 & 62.4 & 77.6 & 94.1 & \cellcolor{green!20}94.1 & 78.8 & 75.1 \\
Qwen-2.5-7B-It & 89.4 & 94.1 & 95.3 & 98.8 & 67.1 & \cellcolor{green!20}90.6 & \cellcolor{green!20}77.6 & \cellcolor{green!20}76.5 & 84.7 & 98.8 & 83.5 & 92.9 & 87.4 \\
Qwen-2.5-72B-It & 74.1 & 70.6 & 90.6 & \cellcolor{green!20}100.0 & 57.6 & 83.5 & 64.7 & 61.2 & 78.8 & \cellcolor{green!20}100.0 & 83.5 & 82.4 & 78.9 \\
Llama-3.1-8B-It & \cellcolor{red!20}28.2 & \cellcolor{red!20}43.5 & \cellcolor{red!20}41.2 & \cellcolor{red!20}72.9 & 37.6 & \cellcolor{red!20}69.4 & \cellcolor{red!20}3.5 & \cellcolor{red!20}60.0 & \cellcolor{red!20}72.9 & \cellcolor{red!20}69.4 & \cellcolor{red!20}67.1 & 75.3 & \cellcolor{red!20}53.4 \\
Llama-3.1-70B-It & 77.6 & 76.5 & \cellcolor{green!20}97.6 & 97.6 & \cellcolor{green!20}85.9 & 77.6 & 76.5 & 62.4 & 80.0 & 96.5 & 72.9 & 96.5 & 83.1 \\
DeepSeek-V3 & 54.1 & 61.2 & 60.0 & 92.9 & \cellcolor{red!20}16.5 & 76.5 & 25.9 & 61.2 & 76.5 & 98.8 & \cellcolor{green!20}94.1 & \cellcolor{red!20}58.8 & 64.7 \\
DeepSeek-R1 & \cellcolor{green!20}91.8 & 81.2 & 88.2 & \cellcolor{green!20}100.0 & 70.6 & 87.1 & 72.9 & 65.9 & 81.2 & 98.8 & 89.4 & 88.2 & 84.6 \\

\midrule
\rowcolor{orange!10} \multicolumn{14}{c}{\textit{\textbf{Turn = 2}}} \\ \midrule

GPT-4.1 & 56.5 & 51.6 & 57.6 & 72.4 & 34.9 & \cellcolor{green!20}64.9 & 44.1 & 47.5 & 57.8 & 68.0 & 56.3 & 71.6 & 56.9 \\

o3 & 49.0 & 35.5 & \cellcolor{red!20}28.0 & 69.0 & 14.7 & 54.9 & 42.5 & 43.3 & 55.1 & 67.3 & 51.8 & \cellcolor{red!20}42.5 & 46.1 \\

Gemini-2.5-Flash & \cellcolor{green!20}73.7 & \cellcolor{green!20}68.2 & 67.3 & 67.6 & 64.5 & 63.7 & 63.7 & \cellcolor{green!20}49.8 & 59.8 & 59.2 & 62.0 & 73.9 & \cellcolor{green!20}64.4 \\

Claude-Sonnet-4.5 & 66.1 & 53.7 & 53.1 & 72.5 & 23.3 & 54.5 & 58.8 & 42.9 & 52.5 & 63.3 & 70.0 & 63.3 & 56.3 \\

Qwen-2.5-7B-It & 50.0 & 58.4 & 52.7 & 63.3 & 40.4 & \cellcolor{red!20}50.8 & 43.5 & \cellcolor{red!20}41.8 & \cellcolor{red!20}46.5 & 53.9 & 53.3 & 53.1 & 50.6 \\

Qwen-2.5-72B-It & 52.4 & 60.8 & 70.8 & \cellcolor{green!20}77.6 & 36.7 & 61.8 & 44.3 & 45.9 & \cellcolor{green!20}61.0 & \cellcolor{green!20}70.2 & 66.1 & 59.8 & 59.0 \\

Llama-3.1-8B-It & \cellcolor{red!20}17.6 & \cellcolor{red!20}29.8 & 28.6 & \cellcolor{red!20}52.0 & 27.3 & 51.6 & \cellcolor{red!20}2.5 & 43.3 & 53.7 & \cellcolor{red!20}47.8 & \cellcolor{red!20}46.5 & 55.5 & \cellcolor{red!20}38.0 \\

Llama-3.1-70B-It & 60.6 & 57.8 & \cellcolor{green!20}76.1 & 73.9 & \cellcolor{green!20}64.7 & 56.3 & \cellcolor{green!20}66.3 & \cellcolor{green!20}49.8 & 59.4 & 67.5 & 57.3 & \cellcolor{green!20}80.0 & 64.2 \\

DeepSeek-V3 & 42.2 & 47.3 & 41.4 & 66.7 & \cellcolor{red!20}11.0 & 53.9 & 20.0 & 43.9 & 52.0 & 66.9 & \cellcolor{green!20}71.6 & 43.5 & 46.7 \\

DeepSeek-R1 & 59.4 & 62.0 & 62.7 & 69.2 & 45.9 & 63.5 & 43.9 & 44.7 & 56.9 & 66.9 & 59.8 & 61.4 & 58.0 \\

\midrule
\rowcolor{purple!9!white} \multicolumn{14}{c}{\textit{\textbf{Turn = 3}}} \\ \midrule

GPT-4.1 & 38.5 & 49.1 & 47.4 & 89.1 & 19.4 & 74.1 & 27.4 & 51.5 & 68.5 & \cellcolor{green!20}94.4 & 49.7 & 61.2 & 55.9 \\
o3 & 20.0 & 33.8 & 24.1 & 65.9 & \cellcolor{red!20}4.4 & 58.2 & 16.8 & 50.0 & 66.5 & 88.5 & 37.9 & 30.9 & 41.4 \\

Gemini-2.5-Flash & \cellcolor{green!20}76.2 & \cellcolor{green!20}98.2 & \cellcolor{green!20}85.3 & 91.5 & \cellcolor{green!20}66.8 & \cellcolor{green!20}83.5 & \cellcolor{green!20}64.7 & \cellcolor{green!20}63.5 & 73.8 & 81.8 & 82.6 & \cellcolor{green!20}89.1 & \cellcolor{green!20}79.7 \\

Claude-Sonnet-4.5 & 54.1 & 68.5 & 64.7 & \cellcolor{green!20}92.6 & 21.5 & 71.8 & 46.8 & 59.7 & \cellcolor{green!20}75.9 & 92.9 & 76.8 & 72.1 & 66.5 \\

Qwen-2.5-7B-It & 13.8 & 23.5 & 15.6 & 34.4 & 10.0 & \cellcolor{red!20}21.5 & 12.6 & 18.8 & \cellcolor{red!20}22.1 & \cellcolor{red!20}38.5 & \cellcolor{red!20}24.7 & \cellcolor{red!20}14.7 & \cellcolor{red!20}20.9 \\
Qwen-2.5-72B-It & 22.4 & 36.8 & 46.2 & 55.6 & 12.1 & 46.5 & 16.8 & 29.7 & 42.6 & 63.5 & 53.5 & 22.6 & 37.4 \\
Llama-3.1-8B-It & \cellcolor{red!20}7.6 & \cellcolor{red!20}16.2 & \cellcolor{red!20}7.6 & \cellcolor{red!20}34.1 & 8.2 & 35.6 & \cellcolor{red!20}0.0 & \cellcolor{red!20}17.6 & 32.4 & 51.2 & 33.8 & 19.7 & 22.0 \\
Llama-3.1-70B-It & 52.1 & 57.1 & 58.8 & 84.7 & 37.6 & 60.3 & 45.3 & 55.9 & 69.7 & 88.5 & 62.1 & 67.6 & 61.7 \\

DeepSeek-V3 & 40.6 & 56.8 & 45.3 & 73.8 & 7.9 & 60.3 & 17.9 & 46.5 & 62.6 & 83.8 & \cellcolor{green!20}83.5 & 37.9 & 51.4 \\
DeepSeek-R1 & 41.5 & 59.1 & 53.5 & 82.6 & 30.6 & 66.5 & 22.6 & 46.2 & 60.3 & 81.8 & 55.3 & 59.4 & 55.0 \\

\bottomrule
\end{tabular}
\end{adjustbox}
\end{table*}

\vspace{-0.1in}
\subsection{Main Results}

\subsubsection{Task 1: Multi-Turn Ambiguous Query Clarification}
\label{sec:5.2.1}

% In this task, the model must identify the ambiguity type underlying the user’s original query, using information accumulated over the entire multi-turn interaction.  
% This requires tracking evolving dialogue context, integrating partial or noisy user responses, and distinguishing among the five ambiguity categories described in Section~\ref{sec:clarification-types}.  
We evaluate model performance across user personas, with results shown in Table~\ref{tab:perf_user_type}.
Qwen-2.5-7B-It tends to answer directly regardless of uncertainty, which boosts its scores on {Precise} and {Refusal} cases but causes sharp drops on ambiguity-heavy types where clarification is required.
Across models, most achieve an overall accuracy above 70\%, yet each displays distinct strengths and weaknesses across personas.
Notably, performance on {Refusal}-style inputs is uniformly low, reflecting a widespread tendency toward over-clarification and underscoring the need for better ambiguity-aware alignment.
Moreover, large reasoning-centric models do not consistently outperform instruction-tuned LLMs, suggesting that explicit reasoning traces alone are insufficient for handling ambiguity-driven behaviors.
Instead, robustness appears more closely tied to alignment quality and pragmatic inference than to raw reasoning depth.
Finally, model scale correlates strongly with robustness: larger models consistently outperform their smaller counterparts, indicating that pragmatic reasoning, uncertainty calibration, and referential grounding all benefit substantially from increased capacity.

Table~\ref{tab:main} reports subtype-level results.
Across ambiguity subtypes, we observe three consistent trends.
First, performance declines markedly as the dialogue progresses: while most models perform reasonably well on the first turn, accuracy drops sharply on the second turn and even more steeply on the third, showing that multi-turn clarification compounds reasoning difficulty.
Second, differences between models become more pronounced with increased depth.
Gemini-2.5-Flash is the most robust model, achieving the highest accuracy in most subtypes, whereas smaller models (\textit{e.g.}, Qwen-2.5-7B-It, Llama-3.1-8B-It) degrade severely.
Third, despite strong first-turn accuracy, variability across subtypes remains large.
These findings highlight that ambiguity resolution is fragile under multi-turn interaction and that both model capacity and ambiguity subtype heavily influence robustness.

\textit{Takeaway 1.}
Most LLMs exhibit a strong tendency to under-clarify as dialogue depth increases, particularly for smaller models.  
In addition, robustness varies across user personas, indicating that user behavior is a major factor shaping clarification performance.

\begin{figure}[t]
    \centering
    \vspace{0.1in}
\includegraphics[width=0.48\textwidth]{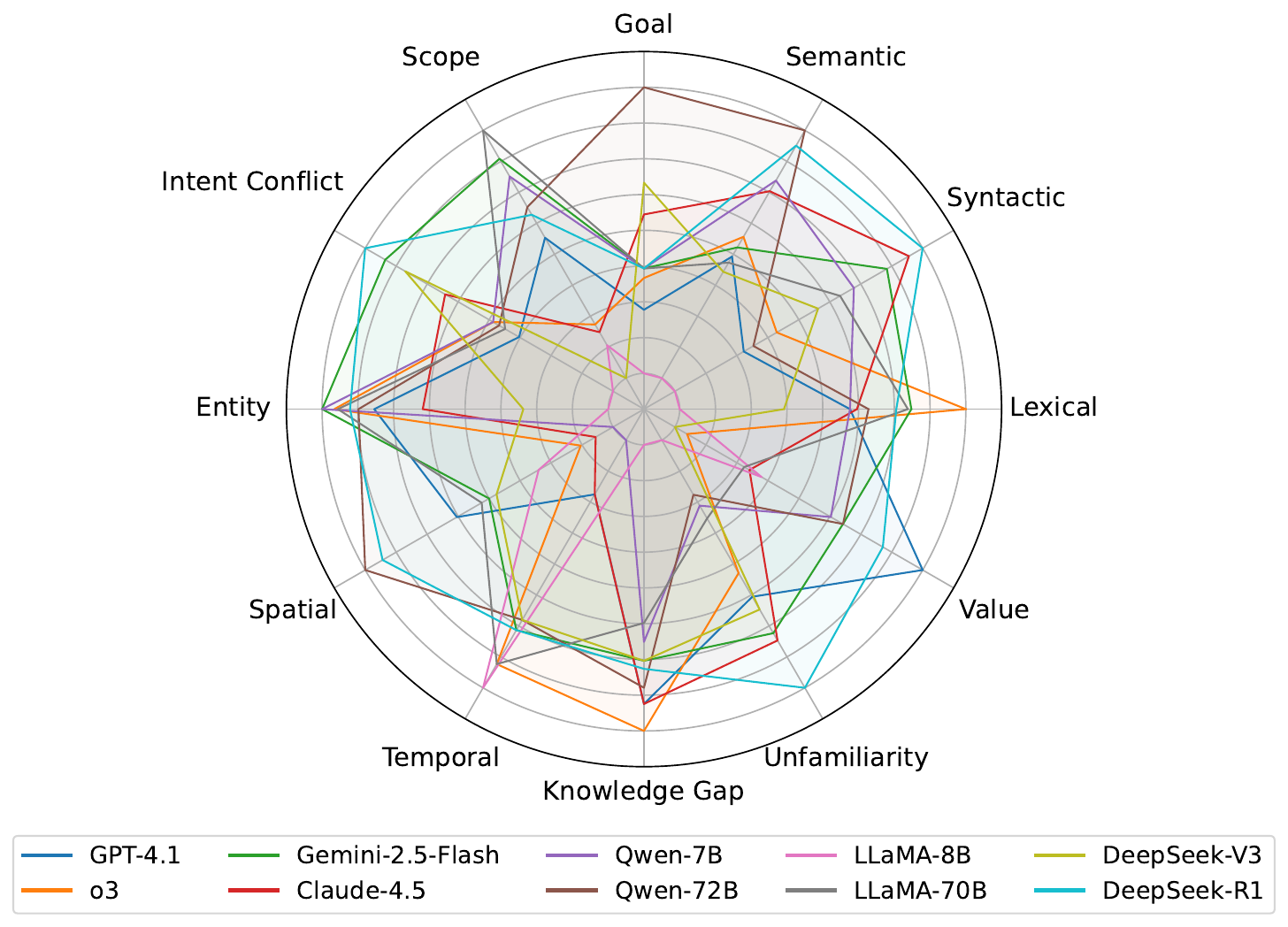}
\vspace{0.01in}
    \caption{
    % Data Source distribution.
    Clarifying question quality for each ambiguity subtype evaluated by LLM-as-a-Judge.
    }
\label{fig:quality}
\end{figure}

\begin{figure}[t]
% \vspace{-0.1in}
\vspace{0.1in}
    \centering
\includegraphics[width=0.48\textwidth]{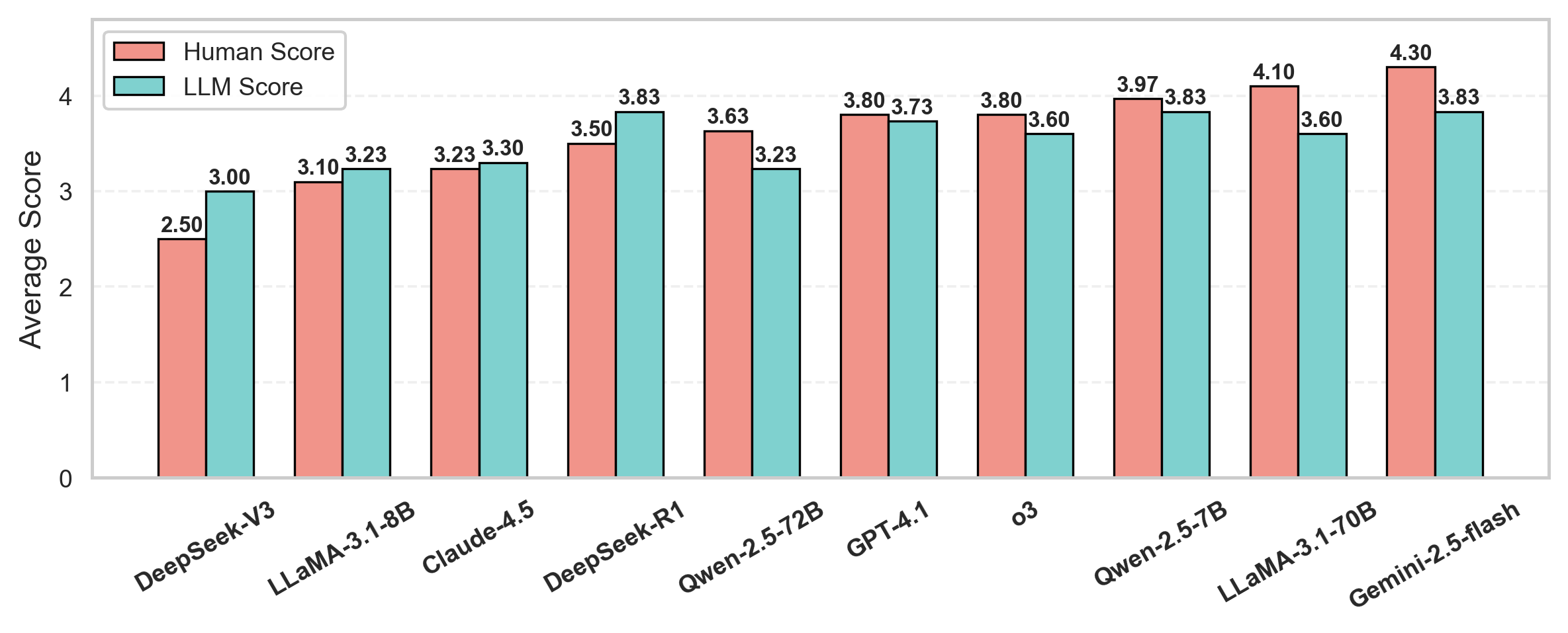}
% \vspace{-0.1in}
\vspace{0.01in}
    \caption{
    % Data Source distribution.
    Clarifying question quality evaluated by human and LLM-as-a-Judge.
    }
\label{fig:human}
\vspace{-0.1in}
\end{figure}

\subsubsection{Task 2: Clarifying Question Quality Evaluation}

We evaluate clarifying question quality using an LLM-as-a-Judge evaluation protocol.
We randomly sample 30 instances from each ambiguity subtype and use GPT-4.1 to assign quality scores.
For each instance, the judge assigns a score between 0 and 5, where 5 indicates a question that is fully appropriate for resolving the underlying ambiguity, and 0 indicates an irrelevant or unhelpful question.
The prompt template is shown in Figure~\ref{fig:prompt_llmjudge}.
% Figure~\ref{fig:quality} presents the model-wise quality scores across ambuguity subtypes.

Figure~\ref{fig:quality} presents model-wise quality scores across ambiguity subtypes.
Across ambiguity subtypes, stronger models consistently produce higher-quality clarifying questions, while smaller models (\textit{e.g.}, Llama-3.1-8B-It) perform substantially worse.
Performance is highest on concrete ambiguity types, such as under-specified goals, missing entities, or attribute-level gaps, where the missing information is explicit.
In contrast, quality degrades on more abstract forms of ambiguity, such as high-level intent or latent preference gaps, where the desired clarification requires implicit reasoning.
Overall, these patterns suggest that current LLMs are much more reliable at resolving explicit missing information than at inferring latent user intent or addressing discourse-level under-specification.

We additionally conduct a human evaluation using the same 0–5 rating rubric.
The results is shown in Figure~\ref{fig:human}. We observe a Pearson correlation of 0.658 between human and LLM scores, indicating moderately strong alignment.
Human ratings tend to be more extreme, exhibiting larger variance across models, whereas LLM scores are more conservative, a pattern consistent with prior reports that LLM judges avoid assigning very high or very low ratings~\cite{li2025generation}.
Both human annotators and the LLM evaluator consistently assign lower scores to models such as DeepSeek-V3 and Llama-3.1-8B-It, while Gemini-2.5-Flash receives the highest average score from both sources, mirroring its strong performance in Table~\ref{tab:main}.
This convergence further validates the reliability of the LLM-based evaluation.

\textit{Takeaway 2.}
Clarifying question quality largely correlates with ambiguous query clarification accuracy. 
Meanwhile, different models exhibit distinct strengths across ambiguity subtypes, with variation in both clarification accuracy and question quality.

\section{ClarifyAgent: An Agentic Method for Multi-Turn Clarification}

\begin{figure}[!t]
\vspace{0.1in}
    \centering
    \includegraphics[width=0.5\textwidth]{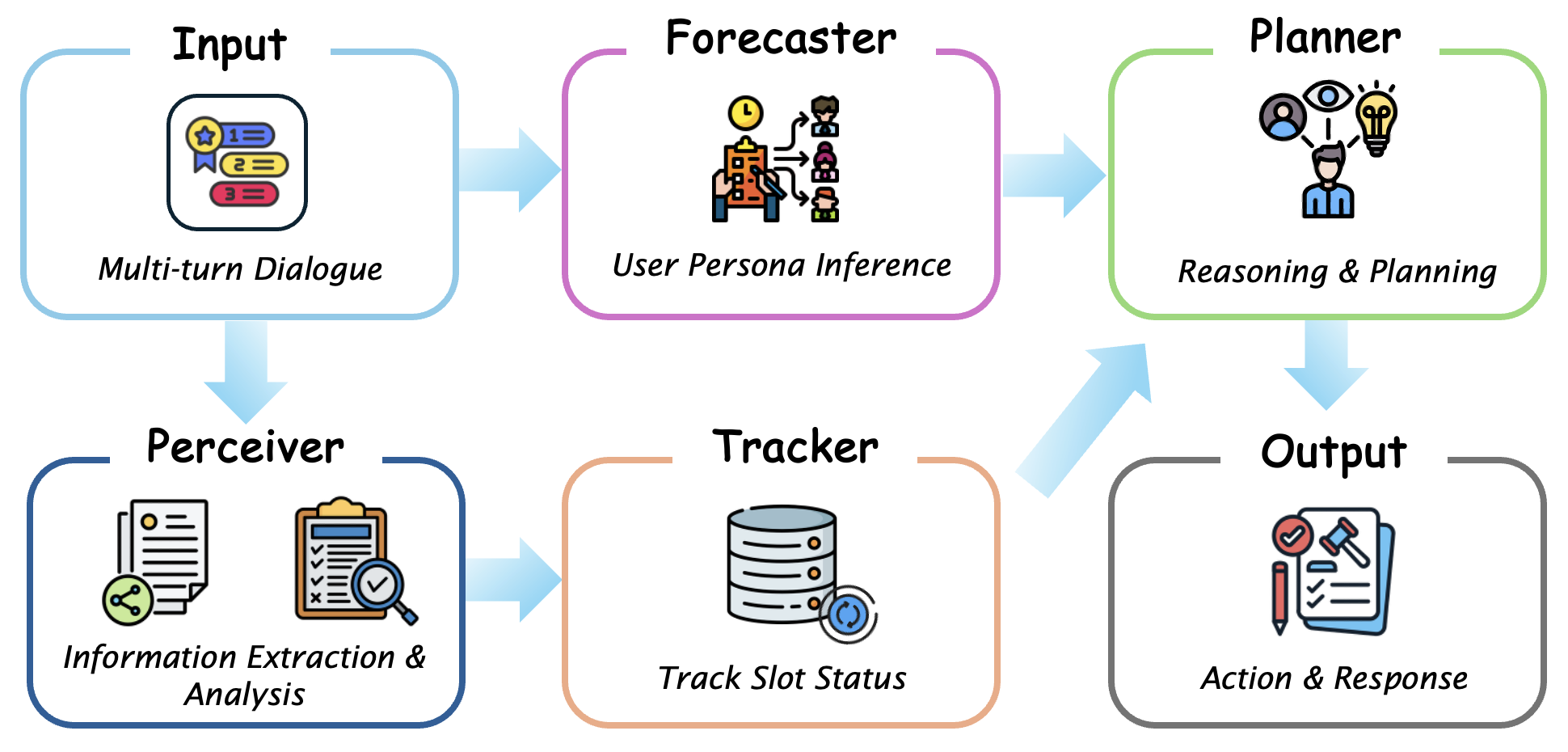}
    
    \caption{Pipeline of ClarifyAgent.}
    \label{fig:method}
    \vspace{-0.1in}
\end{figure}

To improve multi-turn clarification in conversational LLMs, we propose \model, an agentic framework designed to better handle noisy and inconsistent user behavior. We introduce a new subtask, \textbf{user persona inference}, and integrate it into a ReAct-style~\cite{yao2022react} perception--action loop augmented with a finite-state slot tracker. \model{} enables dynamic decision-making about when to ask and when to answer across complex multi-turn interactions. The overall pipeline is illustrated in Figure~\ref{fig:method}.

\vspace{-0.05in}

\subsection{Framework}

\textit{Architecture Overview.}
\model{} is composed of six modular components: an \emph{Input} interface, a \emph{Perceiver} for information extraction, a \emph{Forecaster} for user persona inference, a \emph{Tracker} for finite-state slot management, a \emph{Planner} for reasoning and decision selection, and an \emph{Output} module for action and response generation. Together, these components form a closed-loop perception–reasoning–action cycle that supports multi-turn clarification under noisy or inconsistent user behavior.

At a high level, \model{} executes a goal-directed interaction strategy that tightly couples tracking, forecasting, and planning. At each turn, it parses the user’s latest utterance and updates the Tracker’s slot states; the Forecaster then re-estimates the user persona and cooperativeness based on newly observed evidence. When important information is missing or conflicting, the Planner produces a focused clarification query. When the tracked slots are judged to be sufficiently resolved and the remaining ambiguity is low, the Planner instead synthesizes a task-completing answer.

\textit{Input.}
The Input module receives the multi-turn dialogue and normalizes it into a structured context representation. This representation serves as the entry point for downstream slot extraction, user inferring, and reasoning, ensuring that \model{} maintains a coherent view of the evolving conversation.

\textit{Perceiver.}
Given the current dialogue context, the Perceiver identifies candidate slot values and detects inconsistencies. For each slot, it assigns a discrete status label: \textit{filled}, \textit{unfilled}, or \textit{conflict}—thereby converting raw natural-language input into structured perceptual signals. These signals summarize what is known, missing, or contradictory, and are passed to the Tracker for further reasoning.

\textit{Forecaster.}
The Forecaster estimates the user’s behavioral persona from six categories: {Precise}, {Partial--Vague}, {Off--Focus}, {Contradictory}, {Factually--Wrong}, and {Refusal}.  
It outputs a structured persona label that captures interaction-level uncertainty and conditions the Planner’s ask–answer decisions. This helps avoid unnecessary clarifications with cooperative users and prevents premature answering under noisy or inconsistent behavior.

\textit{Tracker.}
The Tracker maintains the evolving state of ambiguous slots using a finite-state machine (FSM) functioning as a state keeper.  Each slot transitions among \textit{unfilled}, \textit{filled}, and \textit{conflict} as the dialogue unfolds. When all required slots are filled and no conflicts remain, the Tracker notifies the Planner that the \emph{Required Slot Completion (RSC)} condition is satisfied.

\textit{Planner.}
The Planner drives \model{}’s decision process via a perception–reasoning–action loop. At each turn, it integrates the FSM state, the inferred user persona, and the dialogue context to select between clarification and answering. When clarification is needed, the Planner selects the target information to query and instructs the Output module to generate a targeted clarifying question. When the RSC condition is satisfied or further clarification is unlikely to yield additional information, the Planner switches to {Answer} mode and delegates final response generation to the Output module.

% \textit{Output.}
% The Output module surfaces the agent’s final decision (\textit{e.g.}, clarification vs.\ answering) and, when appropriate, an explicit indication of residual uncertainty or conditional assumptions. This explicit handling of uncertainty enhances transparency and provides users with clearer expectations about the reliability of the generated response.

\textit{Output.}
The Output module generates the final action, either a clarifying question or an answer.  
When relevant, it also conveys any remaining uncertainty or conditional assumptions (\textit{e.g.}, “If you mean X, …”).  
By making such uncertainty explicit, the module provides users with clearer expectations about the reliability and scope of the model’s response.

\subsection{Empirical Performance Evaluation}

\begin{table}[!t]
\centering
% \vspace{0.02in}
\caption{Accuracy across six user personas: Precise (P), Partial--Vague (PV), 
Off--Focus (OF), Contradictory (CT), Factually--Wrong (FA), and Refusal (RF). The best results are highlighted in boldface, and * denotes statistical significance at $p < 0.05$ under a paired $t$-test.}
\label{tab:clarifyagent}
\begin{adjustbox}{width=\linewidth}
\begin{tabular}{lccccccc}
\toprule
\textbf{Method} & \textbf{P} & \textbf{PV} & \textbf{OF} & \textbf{CT} & \textbf{FA} & \textbf{RF} & \textbf{Avg.} \\
\midrule

\multicolumn{8}{l}{\textbf{Backbone: Llama-3.1-8B-It}} \\
Base Model        & 48.3 & 75.6 & 87.5 & 86.4 & 84.8 & 44.4 & 71.2 \\
Majority Voting   & 48.3 & 76.9 & 86.4 & 89.0 & 91.2 & 43.3 & 72.4 \\
CoT               & 12.7 & 80.2 & 89.0 & 88.6 & 90.1 & 27.3 & 64.7 \\
Intent-Sim        & 22.6 & 83.3 & 84.4 & 86.6 & 77.6 & 15.0 & 61.6 \\
AT-CoT            & \textbf{74.5} & 80.4 & 84.8 & 81.8 & 87.0 & 29.2 & 73.0 \\
\rowcolor{gray!18} \textbf{\model}      & 58.0 & \textbf{97.2*} & \textbf{100.0*} & \textbf{97.1*} & \textbf{96.5*} & \textbf{81.7*} & \textbf{88.4*} \\

\midrule

\multicolumn{8}{l}{\textbf{Backbone: Qwen-2.5-7B-It}} \\
Base Model        & \textbf{96.6} & 36.5 & 60.3 & 42.4 & 38.8 & 72.9 & 57.9 \\
Majority Voting   & 96.4 & 42.7 & 74.6 & 49.1 & 43.2 & 71.6 & 62.9 \\
CoT               & 89.0 & 55.0 & 63.1 & 65.6 & 62.7 & 63.1 & 66.4 \\
Intent-Sim        & 65.2 & 59.4 & 60.0 & 60.8 & 61.4 & 50.1 & 59.5 \\
AT-CoT            & 93.9 & 73.2 & 69.7 & 54.7 & 50.8 & 43.7 & 64.3 \\
\rowcolor{gray!18} \textbf{\model}            & 85.3 & \textbf{85.8*} & \textbf{97.3*} & \textbf{91.7*} & \textbf{84.0*} & \textbf{83.9*} & \textbf{88.0*} \\

\bottomrule
\end{tabular}
\end{adjustbox}
\end{table}

\begin{table}[!t]
\centering
% \vspace{0.05in}
\caption{Ablation study of \model with Qwen-2.5-7B-It backbone. 
We report the accuracy across six user personas, as well as the average accuracy (Avg.) and the standard deviation (Std.) for each variant.
The best results are highlighted in boldface, and the runner-up results are underlined.
}
\label{tab:agentablation}
\begin{adjustbox}{width=\linewidth}
\begin{tabular}{lcccccccc}
\toprule
\textbf{Variant} & \textbf{P} & \textbf{PV} & \textbf{OF} & \textbf{CT} & \textbf{FA} & \textbf{RF} & \textbf{Avg.} & \textbf{Std.} \\
\midrule
\textbf{\model} & \underline{85.3} & \underline{85.8} & \textbf{97.3} & \textbf{91.7} & \underline{84.0} & {83.9} & \textbf{88.0} & \textbf{4.92} \\
\textit{w/o} Planner & 75.2 & 79.9 & \underline{94.1} & 81.8 & 81.3 & \underline{85.7} & 83.0 & \underline{5.86} \\
\textit{w/o} Perceiver & 62.0 & \textbf{91.6} & 86.2 & \underline{89.1} & \textbf{85.2} & 78.7 & \underline{82.1} & 9.85 \\
\textit{w/o} Forecaster & \textbf{91.8} & 67.2 & 90.9 & 72.1 & 75.2 & \textbf{91.9} & 81.5 & 10.28 \\
\bottomrule
\end{tabular}
\end{adjustbox}
% \vspace{0.05in}
\end{table}

\textit{Evaluation Setup.}
We evaluate \model on ClarifyMT-Bench using two representative open-source base LLMs: Llama-3.1-8B-Instruct and Qwen-2.5-7B-Instruct.
We compare against several training-free baselines, including Majority Voting~\cite{wangself}, Chain-of-Thought (CoT)~\cite{wei2022chain}, Intent-Sim~\cite{zhang2025clarify}, and AT-CoT~\cite{tang2025clarifying}.
All methods operate on the same underlying backbone and are evaluated on the second-turn dialogue for broad coverage of six user personas.
Specifically, Majority Voting uses a decoding temperature of 0.7 and aggregates predictions over $k{=}5$ independent samples.
CoT follows a standard zero-shot chain-of-thought prompting setup.
Intent-Sim also relies on $k{=}5$ samples, with a similarity threshold $\tau=0.5$  to determine whether to query the user.
Moreover, AT-CoT is adapted by slightly modifying its prompt templates to ensure appropriate ask-or-answer decisions in the multi-turn clarification setting.

\textit{Analysis of Effectiveness and Efficiency.}
Table~\ref{tab:clarifyagent} reports accuracy for each user persona. 
Across both backbones, \model{} consistently outperforms all prompting-based baselines by a notable margin.
With Llama-3.1-8B-Instruct, \model{} achieves an average accuracy of 88.4\%, outperforming the strongest baseline by 15.4 absolute points.
The improvements are especially pronounced on noisy personas such as Partial--Vague, Off--Focus, Contradictory, and Factually--Wrong.
A similar trend is observed on Qwen-2.5-7B-Instruct, where \model{} improves the average accuracy from 66.4\% to 88.0\%.
Despite these gains, \model{} does not explicitly optimize for the Precise persona.
Instead, it prioritizes balanced and robust behavior across all six personas, achieving the highest average accuracy among compared methods while remaining competitive on precise inputs.
This trade-off is desirable in multi-turn clarification, where real-world failures often arise from ambiguous or strategically unhelpful feedback rather than fully specified user queries.
In terms of efficiency, \model{} requires five forward passes of the base LLM (once per module), matching the inference cost of Majority Voting and Intent-Sim, both of which also rely on $k{=}5$ samples.
Under this comparable computational budget, \model{} delivers substantially larger accuracy gains, indicating that its performance improvements stem from more effective decision-making rather than increased inference cost.

\textit{Ablation Study.}
Table~\ref{tab:agentablation} presents an ablation study on the Qwen-2.5-7B-Instruct backbone, where we systematically disable each module of \model.
We observe that removing any component leads to a noticeable drop in overall accuracy.
In addition, the standard deviation across personas nearly doubles when either the Perceiver or Forecaster is removed, suggesting that these modules are essential for maintaining balanced performance across heterogeneous user behaviors rather than overfitting to a particular subset.
Each module also contributes distinct behavioral effects.
For instance, eliminating the Forecaster causes the model to become very strong on Precise and Refusal personas, while its performance degrades markedly on other personas.
This indicates that user persona inference is crucial for preserving robustness under noisy or ambiguous feedback.

Overall, the results demonstrate that \model{} delivers pronounced accuracy gains over strong baselines, remains robust under diverse and noisy user behaviors, and achieves a well-calibrated balance between answering and clarification.
The ablations further show that all modules contribute to both accuracy and stability across personas, highlighting their complementary roles within the agentic framework.

% \vspace{-0.1in}

\section{Conclusion}

We presented \textbf{\dataset}, a benchmark for evaluating multi-turn clarification in open-domain human–LLM interactions. Grounded in a five-dimensional ambiguity taxonomy and six behaviorally diverse user personas, \dataset{} enables controlled evaluation of when an LLM should ask, what it should ask, and when it should stop asking. Using this framework, we conducted a systematic study across ten off-the-shelf LLMs and uncovered a consistent {under-clarification bias}: models tend to answer prematurely, struggle to remain robust under noisy or contradictory user feedback, and degrade substantially as dialogue depth increases.
To address this gap, we proposed \textbf{\model}, an agentic approach  that decomposes clarification into perception, forecasting, tracking, and planning. ClarifyAgent significantly improves ask–answer decisions across user personas, offering a strong baseline for future work.
Together, \dataset{} and \model{} reveal key limitations in current conversational LLMs and provide a basis for developing safer, more reliable, and more uncertainty-aware multi-turn interaction.

% Looking forward, ClarifyMT-Bench opens several avenues for future research, including learning-based clarification policies, uncertainty-aware dialogue management, and broader investigations into safety, trust, and user experience in multi-turn human–AI interaction. 

\bibliographystyle{ACM-Reference-Format}
\bibliography{ref}

\newpage

$ $

\newpage

\appendix

\begin{figure}[t]
\centering
\begin{tcolorbox}[
colback=cyan!4,
colframe=cyan!40!black,
  % boxrule=0.5pt,
  % arc=2pt,
  width=\linewidth,
  % boxrule=0.4pt,
  % left=4pt,right=4pt,top=6pt,bottom=6pt
  ]
\textbf{Prompt Template for LLM-as-a-Judge}

You are an impartial evaluation assistant.
You will be provided with a candidate question, a golden response (the ideal answer), and a candidate response (the model’s answer to evaluate).
Your task is to assign an alignment score indicating how well the candidate response matches the golden response.

{Scoring guidelines (0--5):}
\begin{itemize}[leftmargin=1.2em]
    \item \textbf{5} — Fully aligned. Semantically equivalent; no missing key information; no contradictions.
    \item \textbf{4} — Mostly aligned. Minor omissions or differences, but overall meaning preserved.
    \item \textbf{3} — Partially aligned. Contains some correct elements but lacks important information.
    \item \textbf{2} — Weak alignment. Only small portions match; the majority is incomplete or off-target.
    \item \textbf{1} — Barely aligned. Very limited semantic overlap.
    \item \textbf{0} — Not aligned. Irrelevant, contradictory, or does not follow the intent of the golden response (e.g., golden response asks for clarification, but the candidate provides a direct answer).
\end{itemize}

Instructions:
\begin{itemize}[leftmargin=1.2em]
    \item Judge semantic meaning, not surface wording.
    \item Be strict: only credit what the golden response explicitly or implicitly contains.
    \item Output only a single integer from 0 to 5 with no additional explanation.
\end{itemize}

Inputs:
Candidate question: <question>\\
Golden response: <golden\_clarifying>\\
Candidate response: <candidate\_response>

\end{tcolorbox}

\caption{Prompt template for LLM-as-a-Judge evaluation.}
\label{fig:prompt_llmjudge}
\end{figure}

\section{Ethical Statement }

This work does not involve the collection or analysis of real user data. All dialogues in ClarifyMT-Bench are synthetically generated by LLMs or collected from open-source dataset. They are manually validated, and therefore contain no personal, identifiable, or sensitive information. The benchmark poses no privacy or data-protection risks, and does not target any demographic group. Although the dataset models challenging behaviors such as contradictory or factually incorrect replies, these behaviors are abstracted templates rather than reflections of real individuals. Overall, this study introduces minimal ethical risk and aligns with responsible research practices for human–LLM interaction.

\section{Data Source Distribution}
\label{sec:adddata}

\begin{figure}[t]
    \centering
\includegraphics[width=0.32\textwidth]{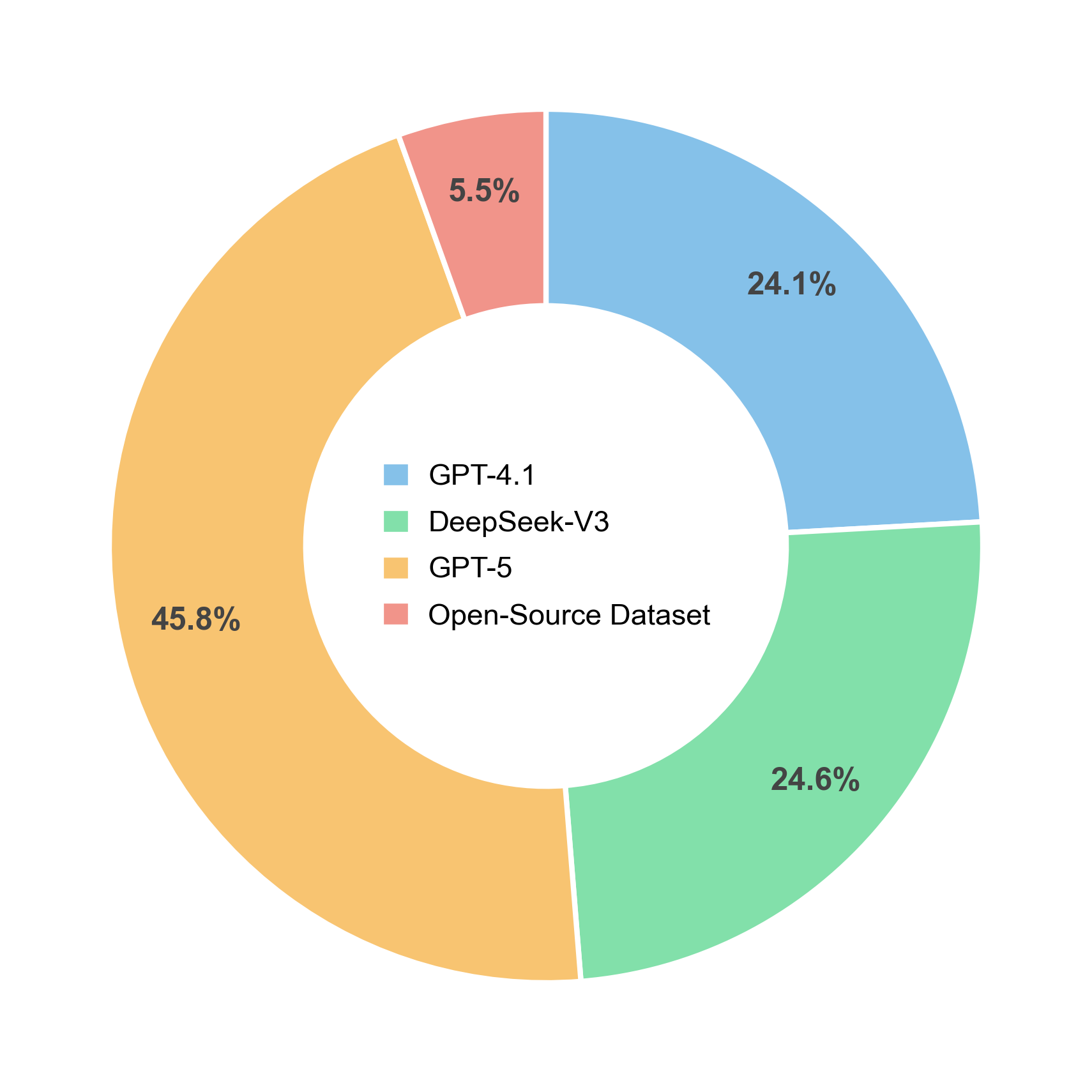}
\vspace{-0.1in}
    \caption{
    % Data Source distribution.
    Distribution of data sources.
The dataset is constructed from diverse sources, primarily leveraging high-capability models including GPT-5, DeepSeek-V3, and GPT-4.1, complemented by open-source datasets to ensure high-quality and diverse coverage.
    }
\label{fig:data_dist}

\end{figure}

We report the data source distribution of our dataset in Figure~\ref{fig:data_dist}. As shown, GPT-4.1 and DeepSeek-V3 each contribute roughly one quarter of the samples, open-source dataset (\textit{i.e.}, CLAMBER) account for about 5\%, and GPT-5 comprises the remaining 45\%. This balanced mixture ensures broad topic diversity and mitigates overfitting to any single model’s conversational style.

% \section{Implementation Details}
% \label{sec:impl}

% Unless otherwise stated, we set the decoding temperature to 0, \textit{i.e.}, greedy decoding.
% Due to resource constraints, Qwen-2.5-7B-Instruct and Llama-3.1-8B-Instruct are locally deployed and served via the vLLM framework \cite{kwon2023efficient}.
% OpenAI models are accessed through their official APIs, while the DeepSeek models are accessed through the \texttt{Bailian} platform.\footnote{\url{https://aliyun.com/product/bailian}}
% All other models are accessed via the \texttt{OpenRouter} API service.\footnote{\url{https://openrouter.ai/}}
% Prompt templates follow the general format used in prior clarification benchmarks~\cite{zhang2024clamber,gan2024clarq}, with minor adjustments to ensure consistency across different model families.
% The complete prompt templates and few-shot exemplars are provided in Figure~\ref{fig:prompt-template}.

\begin{table*}[t]
\centering
\caption{Examples of user responses across six user personas in the travel itinerary scenario.}
\label{tab:user-types-travel}
\begin{adjustbox}{width=\linewidth}
\begin{tabular}{p{2.4cm}|p{8.3cm}|p{10.4cm}}
\toprule
\multicolumn{3}{l}{
\textbf{Scenario} \; Q1: Can you plan a 3-day trip for me?} \\
\multicolumn{3}{l}{
A1 (Clarification): Could you share the destination, dates, budget, or any preferred activities?} \\
\midrule
\textbf{User Persona} & \textbf{Description} & \textbf{Example reply (Q2)} \\
\midrule

\textit{Precise} &
Provides the missing details clearly and specifically &
Kyoto from March 12–14. Budget around \$800. I would like to visit temples. \\
\midrule

\textit{Partial–Vague} &
Partially related but vague, not decidable  &
Somewhere in Japan in early spring.  \\
\midrule

\textit{Off–Focus} &
Responds with information unrelated to the clarification request &
Is Japan safe to travel right now? \\
\midrule

\textit{Contradictory} &
Provides statements that conflict with each other &
Let's keep it cheap. Actually, no need to worry about the budget, make it fancy. \\
\midrule

\textit{Factually–Wrong} &
Provides specific but factually incorrect information &
Plan something in Kyoto near the Eiffel Tower. I want to visit it on the first day. \\
\midrule

\textit{Refusal} &
Declines to clarify and pushes the system to decide &
Just make the plan for me. Anything is fine. \\
\bottomrule
\end{tabular}
\end{adjustbox}
\end{table*}

\begin{table*}[t]
\centering
\caption{Ambiguity taxonomy with representative \textbf{clarifying questions}, illustrating how LLMs can engage in pragmatic disambiguation across linguistic, intentional, contextual, epistemic, and interactional dimensions.}
\small
\renewcommand{\arraystretch}{1.25}
\resizebox{\textwidth}{!}{%
\begin{tabular}{p{3.5cm} p{6.5cm} p{10cm}}
\toprule
\textbf{Subtype} & \textbf{Example} & \textbf{Clarifying Question} \\
\midrule

\textit{Lexical Ambiguity} & Please tell me about the seal. & Do you mean the marine animal or the official stamp? \\

\textit{Syntactic Ambiguity} & List movies from the 1990s starring actors from Canada. & Do you mean 1990s films featuring Canadian actors, or Canadian films from that period? \\

\textit{Semantic Ambiguity} & Is New York the largest city? & In what respect—by population size, land area, or economic output? \\

\midrule

\textit{Goal Ambiguity} & Help me write a report. & Could you specify the type or purpose of the report—academic, professional, or personal? \\

\textit{Scope Ambiguity} & Tell me about quantum computing. & Are you looking for a high-level overview or something more technical? \\

\textit{Intent Conflict Ambiguity} & Summarize \textit{War and Peace} without omitting anything. & Should I prioritize completeness or conciseness in the summary? \\

\midrule

\textit{Entity Ambiguity} & Who is the real Spider-Man? & Are you referring to the comic-book character, a film portrayal, or a real person? \\

\textit{Spatial Ambiguity} & Tell me how to reach London. & From which location are you starting your journey? \\

\textit{Temporal Ambiguity} & When does the meeting start? & Are you referring to today’s meeting or a future one? \\

\midrule

\textit{Knowledge Gap Ambiguity} & You remember the new update, right? & Could you clarify which update you are referring to? \\

\textit{Unfamiliarity Ambiguity} & Find the price of the Samsung Chromecast. & Did you mean a Samsung streaming device or Google’s Chromecast? \\

\textit{Value Ambiguity} & Recommend a good movie. & What criteria define “good” for you—genre, popularity, or critical acclaim? \\

\midrule

\textit{Partial / Vague Reply} & Sort of, I guess. & Could you indicate which part you agree with or find uncertain? \\

\textit{Factually Incorrect Reply} & Paris is the capital of Germany. & Did you mean Paris, France, or perhaps Berlin, Germany? \\

\textit{Contradictory Reply} & It’s urgent. No rush actually. & Should I treat this as a high- or low-priority request? \\

\textit{Off-focus Reply} & Let’s talk about something else. & Of course — would you like to switch to a new topic or pause this discussion? \\

\bottomrule
\end{tabular}
}
\label{tab:ambiguity-clarifications-academic}
\end{table*}

\begin{figure}[t]
\centering
\begin{tcolorbox}[
colback=purple!4!white,
colframe=purple!35!black,
  % boxrule=0.5pt,
  % arc=2pt,
  width=\linewidth,
  % boxrule=0.4pt,
  % left=4pt,right=4pt,top=6pt,bottom=6pt
  ]
\textbf{(a) Prompt Template for Single-turn Dialogue Generation}

You are an expert dialogue designer generating short user queries that exhibit <ambiguity\_type>.

Task:
Produce <number> natural user utterances whose <ambiguity\_description>.

For each example, output:
\begin{itemize}[leftmargin=1.2em]
\item "question": the ambiguous user request
\item "clarifying\_question": a natural follow-up asking what scope the user intends
\item "explanation": explanation of ambiguity
\item Format: JSON per line
\end{itemize}

\rule{\linewidth}{0.4pt}

\textbf{(b) Prompt Template for Multi-turn Dialogue Generation}

Your task is to continue a short dialogue between a user and an assistant. Given a dialogue where the assistant asks a clarifying question, generate how six different types of users would respond in the next turns. Each response must sound natural.  Single-turn types produce one user reply; multi-turn types follow a Q1→A1→Q2→A2 pattern.
Output a JSON object with the following keys: <user\_personas>

Example: <example>

Now continue for the following dialogue:
Q: <user\_query>
A: <assistant\_query>

% \textbf{Requirements}
% \begin{itemize}[leftmargin=1.2em]
% \item Produce one clarification statement.
% \item Avoid labels or meta-commentary.
% \item Be natural and conversational.
% \item Focus on concrete and checkable <user\_criteria>.
% \end{itemize}

\rule{\linewidth}{0.4pt}

\textbf{(c) Prompt Template for Model Evaluation}

You are a helpful conversational assistant. In each turn, given the previous dialogue and the user's latest message, your task is to decide whether to answer the user directly or ask a clarifying question.
If the user's request is clear and specific, respond with the final answer.
If the request remains ambiguous, underspecified, or missing essential information, respond with an appropriate clarifying question instead.

Your response MUST begin with either 'The answer is' or 'The clarifying question is'.

\end{tcolorbox}

\caption{Prompt templates used for (a) single-turn ambiguity generation, 
(b) multi-turn dialogue construction across six user personas, 
and (c) model ask--or--answer evaluation.}
\label{fig:prompt}
\end{figure}

\section{Implementation Details}
\label{sec:impl}
Unless otherwise stated, we set the decoding temperature to 0, \textit{i.e.}, greedy decoding.
Due to resource constraints, Qwen-2.5-7B-Instruct and Llama-3.1-8B-Instruct are locally deployed and served via the vLLM framework \cite{kwon2023efficient}.
OpenAI models are accessed through their official APIs, while the DeepSeek models are accessed through the \texttt{Bailian} platform.\footnote{\url{https://aliyun.com/product/bailian}}
All other models are accessed via the \texttt{OpenRouter} API service.\footnote{\url{https://openrouter.ai/}}
Prompt templates follow the general format used in prior clarification benchmarks~\cite{zhang2024clamber,gan2024clarq}, with minor adjustments to ensure consistency across different model families.
The complete prompt template is provided in Figure~\ref{fig:prompt}.

% \section{Social Impact}
% Ambiguous, vague, or internally inconsistent user inputs are common in real-world conversations, yet our findings show that LLMs often respond prematurely, producing misleading or unsafe outputs—an issue that disproportionately affects users with limited digital literacy or domain knowledge. ClarifyMT-Bench enables systematic evaluation of this under-clarification risk, while ClarifyAgent offers a structured mechanism that encourages safer, uncertainty-aware interaction.
% Our work yields several ethical benefits. It reduces harm from overconfident hallucinations by prompting models to clarify intent before answering, particularly under noisy or imprecise user behavior. It also improves accessibility and equity by providing proactive clarification support to users who may struggle to articulate their goals. In high-stakes domains, this clarification-first approach enhances safety by preventing harmful recommendations that could arise from ambiguous queries. Moreover, ClarifyAgent’s explicit reasoning process and slot-tracking framework promote greater transparency, making model decisions more interpretable.
% Taken together, ClarifyMT-Bench and ClarifyAgent advance more responsible and trustworthy LLM-based interactions, supporting the broader goals of Web4Good and the safe deployment of conversational AI.

\begin{figure}[t]
    \centering
\includegraphics[width=0.48\textwidth]{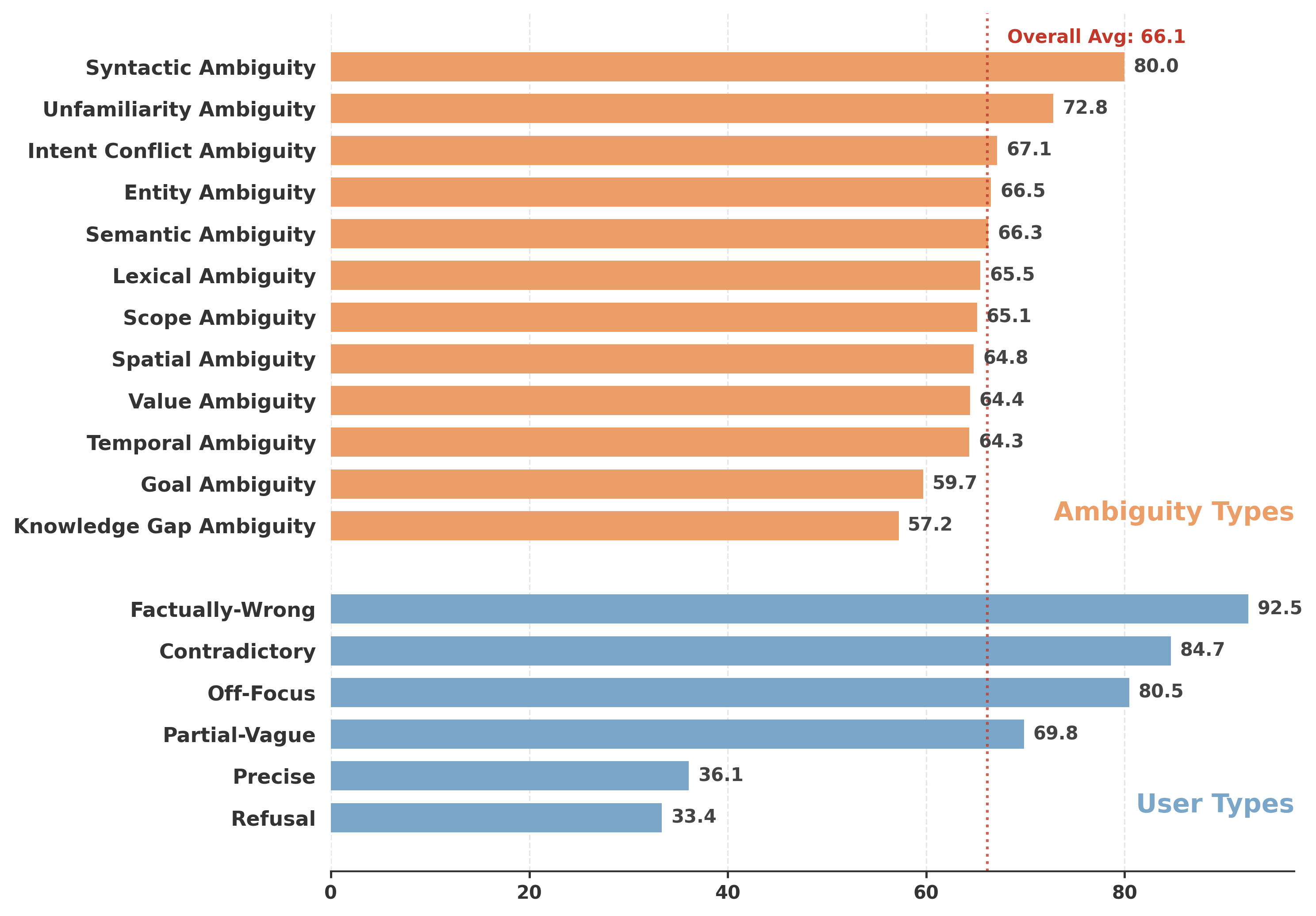}
    \caption{
    Analysis of average dialogue length,   grouped and sorted by ambiguity types (orange) and user types (blue).
    }
\label{fig:dia_length}

\vspace{-0.1in}
\end{figure}

% \end{thebibliography}

\end{document}